\documentclass[conference]{IEEEtran}
\IEEEoverridecommandlockouts
\usepackage{algorithm}
\usepackage{algpseudocode}
\usepackage{cite}
\usepackage{amsmath,amssymb,amsfonts}
\usepackage{graphicx}
\usepackage{textcomp}
\usepackage{xcolor}
\usepackage{comment}
\usepackage{tabularray}
\def\BibTeX{{\rm B\kern-.05em{\sc i\kern-.025em b}\kern-.08em
    T\kern-.1667em\lower.7ex\hbox{E}\kern-.125emX}}
\begin{document}

%Edge Video Compression Using Radiance Fields
\title{
Distributed Radiance Fields for Edge Video Compression and Metaverse Integration in Autonomous Driving
%Distributed Radiance Fields: A Novel Approach for Edge Video Compression and Metaverse Integration in Autonomous Driving
}

\author{\IEEEauthorblockN{Eugen \v{S}lapak\IEEEauthorrefmark{1}, Mat\'{u}\v{s} Dopiriak\IEEEauthorrefmark{1}, Mohammad Abdullah Al Faruque\IEEEauthorrefmark{2}, Juraj Gazda\IEEEauthorrefmark{1}, Marco Levorato\IEEEauthorrefmark{2}} \IEEEauthorblockA{\IEEEauthorrefmark{1}Department of Computers and Informatics, Technical University of Ko\v{s}ice, Slovakia\\ Email: \{eugen.slapak, matus.dopiriak, juraj.gazda\}@tuke.sk} 
\IEEEauthorblockA{\IEEEauthorrefmark{2}Department of Electrical Engineering and Computer Science, University of California, Irvine, United States}
Email: \{alfaruqu, levorato\}@uci.edu
}

\maketitle

\begin{abstract}
 The metaverse is a virtual space that combines physical and digital elements, creating immersive and connected digital worlds. For autonomous mobility, it enables new possibilities with edge computing and digital twins (DTs) that offer virtual prototyping, prediction, and more. DTs can be created with 3D scene reconstruction methods that capture the real world’s geometry, appearance, and dynamics. However, sending data for real-time DT updates in the metaverse, such as camera images and videos from connected autonomous vehicles (CAVs) to edge servers, can increase network congestion, costs, and latency, affecting metaverse services.  Herein, a new method is proposed based on distributed radiance fields (RFs), multi-access edge computing (MEC) network for video compression and metaverse DT updates. RF-based encoder and decoder are used to create and restore  representations of camera images. The method is evaluated on a dataset of camera images from the CARLA simulator. Data savings of up to 80\% were achieved for H.264 \emph{I-frame - P-frame} pairs by using RFs instead of \emph{I-frames}, while maintaining high peak signal-to-noise ratio (PSNR) and structural similarity index measure (SSIM) qualitative metrics for the reconstructed images. Possible uses and challenges for the metaverse and autonomous mobility are also discussed.
\end{abstract}

\begin{IEEEkeywords}
autonomous driving, digital twin, edge computing, metaverse, radiance fields, video compression.
\end{IEEEkeywords}

\section{Introduction}

The advent of connected autonomous vehicles (CAVs) with their supporting technologies enables interactions between vehicles, servers, and peripheral devices. Such technologies include high-resolution CAV sensors, multi-access edge computing (MEC) networks, and metaverse for autonomous mobility, all generating high volumes of traffic. MEC increases the data traffic by transferring the data and computational load from CAVs to MEC servers, while the metaverse uses large amounts of offloaded data to rebuild the real world in a simulated environment, updated in real-time. This interaction mandates the transmission of substantial volumes of data generated by a diverse set of sensors, including cameras, LiDAR, and radar. The hierarchical structure of MEC, with distributed local offloading of data and computations of spatially close CAVs helps to mitigate excessive data transfer, minimizing the necessity for large-scale data transmission to centralized servers. However, the application of advanced data compression techniques is still needed to further decrease the transmission latency. According to the research of Hirlay Alves et al. \cite{alves2022beyond}, enabling the digital twin (DT) in smart city necessitates a network latency in the $5$ to $10$ ms range, and a reliability of $1 - 10^{-5} \%$.

 DTs are virtual models of physical entities, updated in real-time to mirror their real-world counterparts, facilitating design, testing, and optimization in virtual settings, thus improving reliability and reducing costs \cite{glaessgen2012thedigital}. The metaverse, an interactive realm combining physical and digital elements, extends the DT concept by allowing multiple models to co-exist and interact, providing a dynamic platform for autonomous mobility applications \cite{minrui2023full}. This integration enables real-time synchronization with the physical world and the ability to provide immediate feedback for intelligent control of CAVs, leveraging insights from predictive models.

The current generation of advanced machine learning models is capable of learning, reasoning and perform system control based on detailed visual information. Seamless deployment of machine learning models trained in hand crafted simulation to real world is, however, hampered by a gap of graphical fidelity between simulated and real world graphics. Therefore, another aim of the metaverse is to construct photorealistic copies of specific areas, such as city segments, including any dynamic elements. Recent research \cite{alves2022beyond, mihai2022digital, minrui2023full, ren2022quantum, xu2023generative, zhou2023vetaverse} also underscores the importance of implementing metaverse and DTs for reconstructing alternative digital environments, which do not necessitate advanced sensor data, e.g., LiDARs and their subsequent transmission over the network.
 
In this work, we present a solution for the aforementioned problems. The methods presented in our work are capable of rapidly decreasing the amount of transferred data needed to update the metaverse in real time, simultaneously decreasing the latency. Moreover, our approach ensures the lowest possible gap between real-world and simulated visual input and MEC-deployed distributed architecture.

Our solution will use visual data compression employing advances in implicit 3D scene reconstruction, specifically, radiance fields (RFs). RF is a 3D scene representation that can be created just using a sparse set of 2D images. Following the training phase, RF is able to reconstruct any camera view of the scene, including views that are not present in the training set. By equipping the receiver of an image frame from real 3D scene with its RF representation, it would be, in an ideal case, sufficient to send the few bytes of sender's camera pose data to receiver so that it would be able to reconstruct the complete matching sender's view. In practice, however, there are imperfections in scene reconstruction by RF, rapid visual changes due to presence of dynamic objects, changes of scene lighting color and imperfect determination of sender's camera pose, and other factors, all contributing to discrepancies between the actual camera view and RF.

For these reasons, our method equips both sender and the receiver with RF, so that sender can use standard video encoder to encode any differences between the frame rendered from RF and real 3D scene frame. At the receiver end, these differential transformations are reapplied to the rendered RF frame. 
While standard video encoding requires periodic sending of non-differential frames containing the whole view information, our approach can completely omit such frames, and in ideal case decrease the differences between real and virtual 3D scene that need to be encoded. This brings large throughput savings, allows a set of RFs to be used for visually accurate metaverse simulation, and rapid updates of the metaverse, due to faster transmission of such efficiently compressed updates. 

\section{Related Work}

\subsection{Deep Learning-Based Video Compression}
The method presented in \cite{zhang2021implicit} uses implicit neural representation with a separate neural network for each frame, however, individual video frames are represented implicitly instead of the whole 3D scene. The models used in this work are designed to be relatively simple, with their size further decreasing via quantization and use of the similarities between neighboring frames. Additionally, no pretrained network is used on the receiver side.

The study in \cite{chen2022lsvc} introduces two compression schemes, motion residual compression (MRC) and disparity residual compression (DRC), exploiting redundancies in binocular automotive videos. These methods leverage geometric and temporal correlations to compress motion and disparity offsets, using deformable convolution for warping.

Reviews of a wide range of deep learning techniques used for video compression are available in \cite{ding2021advances} and \cite{birman2020overview}.  In \cite{birman2020overview}, the main methods proposed in existing research are categorized as end-to-end schemes, next video frame prediction, generative models and autoencoder schemes. To the best of our knowledge, none of the existing approaches uses encoder and decoder structures based on RFs.

\subsection{RFs as DTs in Autonomous Mobility}

Neural radiance field (NeRF) refers to approximation of 3D scene radiance, capable of reconstructing views on the scene from arbitrary location and angle. Block-NeRF \cite{tancik2022block} demonstrates how neighbourhood-scale NeRF representation can be built from a set of individually trained NeRFs from visually very diverse data collected over timespan of three  months. This work  provides an important proof of practical feasibility of building the distributed set of standalone NeRF neural networks seamlessly modelling large area, that far exceeds volumes manageable by a single NeRF.

Distributed visual data collection for creation of NeRF-based DT for autonomous mobility was examined in \cite{liu2023visualization}. The work has tested the speed of real-world to DT updates when considering different DT quality and network conditions. While this work tests many of the assumptions crucial for our work, it misses the key idea of use of NeRFs for rapid compression to further improve the latency.  

NeRF-based simulator for autonomous driving was developed in \cite{wu2023mars}.
This approach models background environment and foreground objects separately and allows multiple different NeRF backbones and sampling strategies. Modelling the dynamic foreground objects, like vehicles, using NeRFs increases utility of such environment for training of other machine learning models. Limited modification of captured dynamic foreground objects is demonstrated, like addition, deletion, rotation and translation.

Robustness of training other algorithms for downstream tasks with NeRF-based simulators is shown in \cite{byravan2023nerf2real}. Here, robot is trained in NeRF with environment collisions determined by NeRF volume density and fully synthetic dynamic object, as opposed to NeRF rendered ones, with its physics approximating the real one. Policy learned in NeRF-based simulation was successfully transferred into the real world. 

These works show both the proof of potential of large-scale RF-based metaverse with real-time transfer of state from real world into the metaverse, but also the knowledge gap our work tries to close.

%\section{Implicit Representations for 3D scene reconstruction}
\section{Video Compression Using Distributed RFs}

\begin{figure*}[!ht]
  \centering
  \includegraphics[width=1\linewidth]{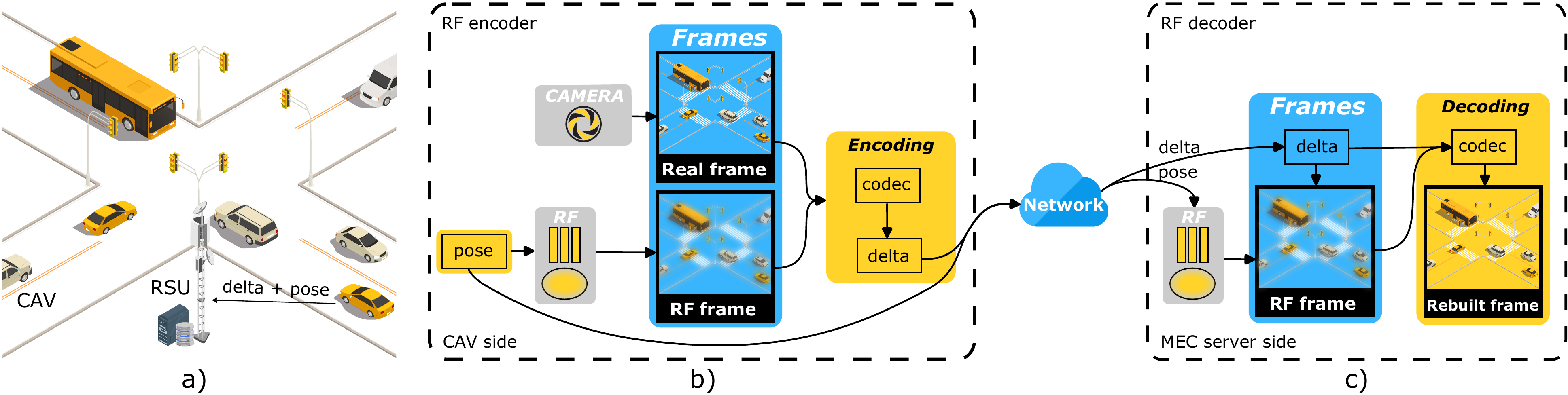}
   %{\epsfig{file = figures/hmap.pdf}}
  \caption{Proposed novel video compression using distributed RFs in MEC network. The diagram shows a) the real-world scene with CAVs and MEC infrastructure b) RF encoder preparing the delta with differences between the real and RF frame c) RF decoder reapplying the delta to RF frame rendered by local copy of the RF.}
  \label{fig:video_compression_rf_av}
 \end{figure*}

\begin{comment}
A reconstruction of 3D photorealistic scenes is a pivotal task to create a digital copy of the real environment, called digital twin. Generally, we can divide 3D geometry representations into two groups, i.e., explicit and implicit representations.
Explicit representation of the surface is defined a function $E$ by the formula:
\begin{equation} \label{eq:explicit_representation}
E :(x, y) \rightarrow z,
\end{equation}
\noindent where $x, y, z \in \mathbb{R}^3$ are the coordinates of the point on the surface and the function $E$ defines all points directly. Meshes, points clouds, voxels belong to the group of 3D geometries based on atomic units as triangles, points and 3D cubes. 

On the other side, implicit representation of the surface is defined as a function $I$ by the formula:
\begin{equation} \label{eq:implicit_representation}
I :(x, y, z) \rightarrow 0,
\end{equation}
\noindent where $x, y, z \in \mathbb{R}^3$ are coordinates of the point on the surface and the function $I$ occupies the particular space and from its nature, we cannot explicitly investigate subsequent points on
the surface. 
\end{comment}

\subsection{Neural Radiance Fields}

NeRFs \cite{mildenhall2021nerf} utilize multi-layer perceptrons (MLPs) to encode 3D scenes into a neural network, parametrizing images with camera poses and optimizing a volumetric scene function approximated by MLP \( F_{\Theta} \):

\begin{equation}
F_{\Theta} :(\boldsymbol{x}, \boldsymbol{d}) \rightarrow (\boldsymbol{\hat{c}}, \sigma),
\end{equation}

\noindent where \( \boldsymbol{x} \) is the location of the point in 3D space, \( \boldsymbol{d} \) is the viewing direction, \( \boldsymbol{\hat{c}} \) is the emitted radiance, and \( \sigma \) is the volume density.

Rendering from RF involves calculating the expected color \( \hat{C}(\boldsymbol{r}) \) of a ray \( \boldsymbol{r}(t) = \boldsymbol{o} + t\boldsymbol{d} \), defined by its origin $\boldsymbol{o}$ and direction $\boldsymbol{d}$, within near and far bounds $t_n$ and $t_f$, by integrating transmittance \( T(t) \): 
%,volume density \( \sigma \), and emitted radiance \( \boldsymbol{\hat{c}} \) along the ray:

\begin{equation}
\hat{C}(\boldsymbol{r}) =\int_{t_n}^{t_f}T(t)\sigma(\boldsymbol{r}(t))\boldsymbol{\hat{c}}(\boldsymbol{r}(t),\boldsymbol{d})dt.
\end{equation}

The loss \( \mathcal{L}_{NeRF} \) is the squared error between rendered $\hat{C}(\boldsymbol{r})$ and ground truth (GT) $C(\boldsymbol{r})$ colors:

\begin{equation}
\mathcal{L}_{NeRF} = \sum_{\boldsymbol{r} \in \mathcal{R}} [\lVert \hat{C}(\boldsymbol{r}) - C(\boldsymbol{r}) \rVert^2_2],
\end{equation}
\noindent where $\mathcal{R}$ is the set of rays in each batch.

Instant neural graphics primitives (INGP) \cite{mullerinstant2022} tackles the issue of NeRFs in excessive training and rendering times using neural graphics primitives and multiresolution hash encoding.
%The model has \( L \) levels, each with up to \( T \) feature vectors of dimension \( F \). Grid resolutions at level \( l \) are determined by:
The model contains trainable weight parameters 
$\phi$ and encoding parameters $\theta$ structured into $L \in \mathbb{N}$ levels, each holding up to $T$ feature vectors of dimension $F$. Each level $l \in L$ operates independently, storing feature vectors at grid vertices. The grid resolution at each level $l$ follows a geometric progression from the coarsest $N_{min}$ to the finest $N_{max}$ resolution by formulas:

\begin{equation}
    N_l := N_\mathrm{min} \cdot b^l,
\end{equation}

\noindent where \( N_l \) is the resolution at level \( l \), and \( b \) is the growth factor:

\begin{equation}
b :=\exp\left(\frac{\ln N_\mathrm{max}-\ln N_\mathrm{min}}{L-1}\right).
\end{equation}

\subsection{3D Gaussian Splatting for RFs}
The 3D Gaussian Splatting (3DGS) \cite{kerbl20233d} uses differentiable 3D Gaussians to model scenes without the use of neural components. 3DGS constructs 3D Gaussians $G(x)$ represented by point (mean) \( \mu \), covariance matrix \( \Sigma \), and opacity \( \alpha \):

\begin{equation}
G(x) = exp(-\frac{1}{2} (x)^T\Sigma^{-1}(x)).
\end{equation}

These Gaussians are projected to 2D using a transformation \( W \) and Jacobian \( J \), resulting in a camera-space covariance matrix \( \Sigma' \):

\begin{equation}
\Sigma' = JW\Sigma W^TJ^T.
\end{equation}

The covariance matrix \( \Sigma \) defines scaling \( S \) and rotation \( R \):

\begin{equation}
\Sigma = RSS^TR^T.
\end{equation}

Stochastic gradient descent (SGD) optimizes Gaussian parameters \( p, \alpha, \Sigma \), and spherical harmonics (SH) representing color $c$ of each Gaussian. The loss function \( \mathcal{L}_{3DGS} \) combines mean absolute error \( \mathcal{L}_{1} \) and a differentiable structural similarity index measure (SSIM) \( \mathcal{L}_{D-SSIM} \) with a balance hyperparameter \( \lambda \):

\begin{equation}
\mathcal{L}_{3DGS} = (1-\lambda)\mathcal{L}_{1} + \lambda\mathcal{L}_{D-SSIM}.
\end{equation}

\subsection{Video Compression Using RFs}
%The proposed novel application of radiance fields for video compression, employing NeRFs, particularly Instant-NGP, and 3D Gaussian Splatting to formulate a digital twin DT. This technique focuses on encoding solely the variances between the DT's anticipated 2D scene representation, as determined by a specific camera pose, and the actual real-time captured frames. This advanced form of video compression significantly diminishes the required transmission bit rate. It achieves this by transmitting only essential data pertaining to camera pose and the differential information between the real and DT-generated frames over the network. An experimental setup was established in a virtual urban environment using the CARLA simulator. Here, we assessed the potential compression savings achieved by an autonomous vehicle when transmitting its video stream utilizing our proposed compression methodology in such a scenario.
%\subsection{Dataset}
Typically, video compression algorithms utilize methods based on optical flow to encode frame differences. However, these methods face constraints in situations where frame changes cannot be adequately described through visual flow transformations of the previous frame, leading to an increased data requirement for each frame. 
%(Fortun et al., 2015).
In stark contrast, our approach uses models with much higher informational content about 3D scenes, surpassing traditional compression by accessing pixel information not available in previously encoded frames. This capability is particularly crucial for capturing details about unseen pixels beyond the camera field of view boundaries or occluded object parts that become visible during movement. \par
%Notably, the storage demands for a trained digital twin's weights are substantially lower than for a single 2D view image of a comparable 3D scene (Mildenhall et al., 2021).
While our approach relies on RFs, real-world camera view will still have multiple differences when compared to RF views, introduced by mobile objects, lighting changes and noise, among other factors. For this reason, we use H.264 compression algorithm to bridge the RF-to-real gap.
H.264 compressed video streams consist of \emph{I-frames} with complete image information, \emph{P-frames} that encode only the differences from the preceding \emph{I-frames}, and \emph{B-frames} capable of encoding differences from temporally bidirectional frames. From now we will refer to the differences encoded by \emph{P-frame} as \emph{P-frame delta}, or just \emph{delta}.
%(Stockhammer, 2003)
Conventionally, every $n$-th frame in a compressed video is an \emph{I-frame}, with parameter $n$ being set prior to video compression. \par
Our proposed approach consists of RF encoder, the encoding data scheme combining camera pose with H.264 encoded difference between real and RF frame and RF decoder as depicted in Fig. \ref{fig:video_compression_rf_av}. RF encoder uses camera pose at CAV to obtain a view rendered by the local copy of the RF and then encodes the difference between real frame and RF rendered one, into \emph{P-frame} delta using H.264 encoder. Camera pose and \emph{P-frame} delta are sent through the wireless network to roadside unit (RSU) that contains local MEC server performing the decoding. RF decoder  decodes the original image from camera pose, used to render the view with RF, and \emph{P-frame} delta, which encodes the differences between the real world and stored RF.\par Both the sender (CAV) and receiver (MEC server) leverage the scene-specific DT to generate and use these images, thus bypassing the need to transmit \emph{I-frames} over the network. \par

The whole process of RF-based encoding, transmission of encoded data, and RF-based decoding is described in pseudocode Alg. 1, using the three corresponding procedures.

\begin{algorithm}
\caption{RF-based Encoder, Network Transmission, and RF-based Decoder}
\begin{algorithmic}[1]

\Procedure{RFEncoder}{$\text{CAV\_Images}, \text{RF}$}
    \State $\text{Encoded\_Frames} \gets \text{empty list}$
    \For{each $\text{image} \in \text{CAV\_Images}$}
        \State $\text{pose} \gets \text{ExtractPose}(\text{image})$
        \State $\text{RF\_frame} \gets \text{RenderViewFromRF}(\text{RF}, \text{pose})$
        \State $\text{delta} \gets \text{EncodeDifference}(\text{image}, \text{RF\_frame})$
        \State $\text{Add}(\text{Encoded\_Frames}, (\text{pose}, \text{delta}))$
    \EndFor
    \State \textbf{return} $\text{Encoded\_Frames}$
\EndProcedure

\\

\Procedure{NetworkTX}{$\text{Encoded\_Frames}, \text{Throughput}$}
    \State $\text{Transmitted\_Frames} \gets \text{empty list}$
    \For{each $(\text{pose}, \text{delta}) \in \text{Encoded\_Frames}$}
        \State $\text{frame\_size} \gets \text{GetSize}(\text{pose}, \text{delta})$
        \State $\tau \gets \text{frame\_size} / \text{Throughput}$
        \State $\text{SendFrameOverNetwork}(\text{pose}, \text{delta}, \tau)$
        \State $\text{Add}(\text{Transmitted\_Frames}, (\text{pose}, \text{delta}, \tau))$
    \EndFor
    \State \textbf{return} $\text{Transmitted\_Frames}$
\EndProcedure

\\

\Procedure{RFDecoder}{$\text{Transmitted\_Frames}, \text{RF}$}
    \State $\text{Decoded\_Images} \gets \text{empty list}$
    \For{each $(\text{pose}, \text{delta}, \tau) \in \text{Transmitted\_Frames}$}
        \State $\text{WaitForTransmission}(\tau)$
        \State $\text{RF\_frame} \gets \text{RenderViewFromRF}(\text{RF}, \text{pose})$
        \State $\text{image} \gets \text{DecodeDifference}(\text{RF\_frame}, \text{delta})$
        \State $\text{Add}(\text{Decoded\_Images}, \text{image})$
    \EndFor
    \State \textbf{return} $\text{Decoded\_Images}$
\EndProcedure
\label{alg:rf_encoder_decoder}
\end{algorithmic}
\end{algorithm}

An experimental setup was established in a virtual urban environment using the CARLA simulator to validate this approach. This setup aimed to replicate real-world urban scenarios within a controlled, simulated environment, providing a robust platform for testing the efficacy of the RF-based compression system in an urban CAV context.\par
%In this scenario, a well-structured network architecture within a Mobile Edge Computing (MEC) network is crucial. Such a network would effectively distribute computational loads, allowing for the offloading of digital twin-related computations from the autonomous vehicle to edge servers, thereby maximizing efficiency and minimizing onboard data processing demands.\par

\subsection{Qualitative Metrics}
To evaluate both the possible quality degradation introduced by RFs and differences between the RFs and changes in the environment (different lighting, presence of vehicles at new locations, etc.) we have used multiple metrics widely applied for image quality comparisons. These include peak signal-to-noise ratio (PSNR), SSIM, and learned perceptual image patch similarity (LPIPS).
These metrics measure the difference between a reference image and a distorted image in terms of pixel values, structural features, and perceptual features, respectively.

\section{Experimental Results}

An experimental framework was established to evaluate the quality of the 3D scene reconstructed by RFs and compression savings of a newly RF-based compression algorithm within an urban 3D environment, as illustrated in Fig. \ref{fig:carla_ir_rl_ingp_gs_gt2}. 
\subsection{Dataset}
\begin{figure*}[!ht]
  \centering
  \includegraphics[width=0.75\linewidth]{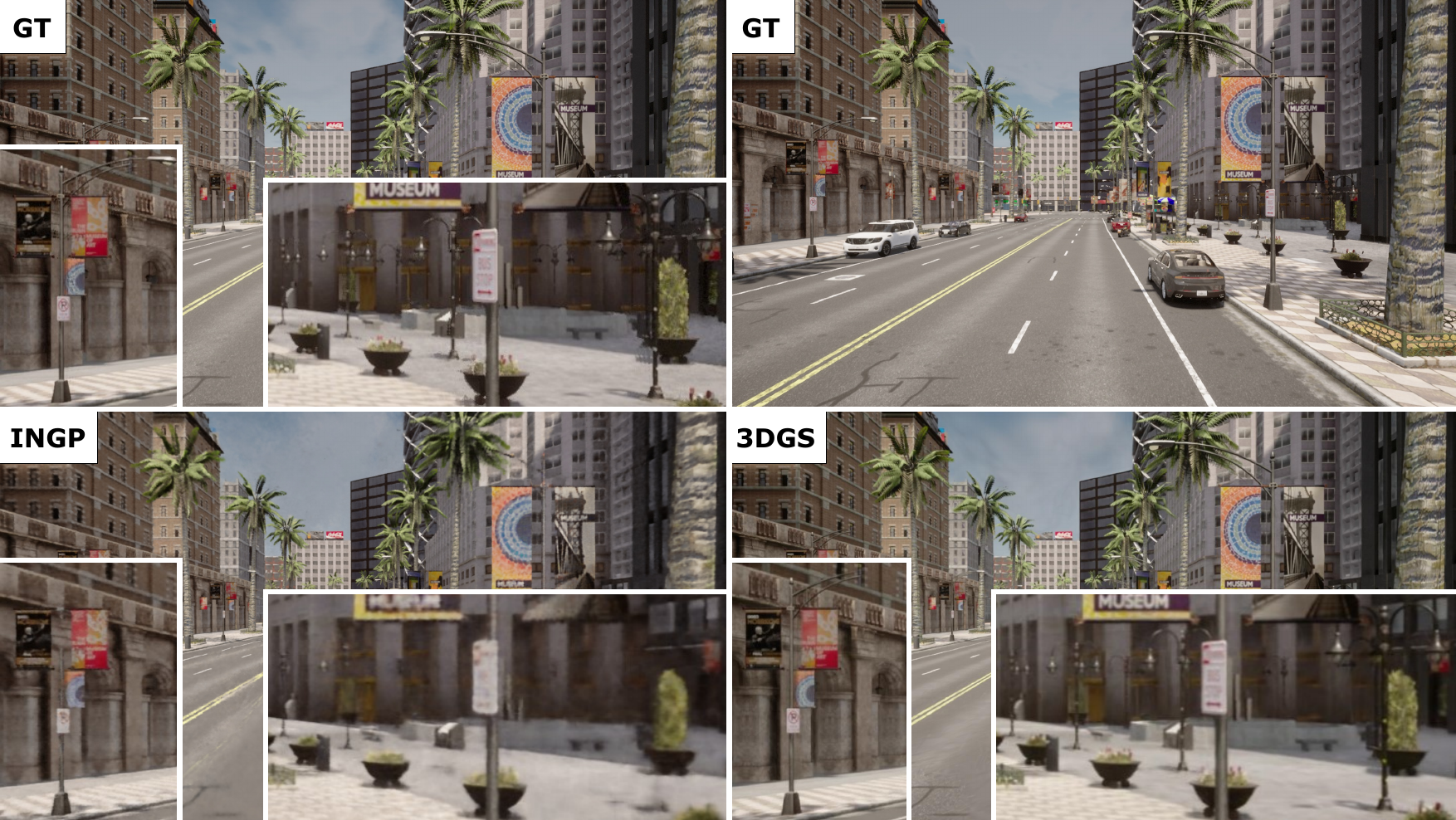}
   %{\epsfig{file = figures/hmap.pdf}}
  \caption{GT images from CARLA simulator with matched INGP and 3DGS images from the same camera pose. Note the varying degree of blur and missing details in INGP and 3DGS rendered frames, like distortion of the letters in the "MUSEUM" sign and missing parts of the lamp structures.}
\label{fig:carla_ir_rl_ingp_gs_gt2}
 \end{figure*}

A training dataset for the RFs was meticulously compiled by rendering the specified urban 3D scene utilizing the CARLA simulation software using 18 cameras attached to the car driving in both directions of the street. Fig. \ref{fig:carla_ir_rl_ingp_gs_gt2} depicts images from CARLA simulator as GT for two scenarios, empty road and parked vehicles on the sides of the road. The compression efficiency of the proposed approach was systematically evaluated by plotting the compression gains across 144 frames captured from multiple cameras positioned along the vehicle's trajectory.\par
\subsection{H.264 Encoding}
Fig. \ref{fig:step_plot_compression_savings_carla_ir_rl_noon_empty_ingp} illustrates the percentages of compression savings achieved at varying resolutions using INGP in the scene without vehicles. The  efficiency of our proposed method is quantified in terms of compression savings, which are determined in relation to a baseline established by frame pairs (\emph{I-frame} and \emph{P-frame}) compressed using the H.264 codec. This assessment encompassed a range of frame resolutions, extending from 300 $\times$ 168 to 1920 $\times$ 1080 (full HD). Images within the dataset were encoded utilizing configuration settings, specifically presets including 'veryslow', 'medium', and 'veryfast', coupled with constant rate factors (CRFs) of 18, 23, and 28. Subsequently, the values obtained for each encoded image pair were averaged to derive the final results. Notably, in our approach, the \emph{I-frame} is not transmitted over the network. Instead, it is generated using RFs. The formula employed for the calculation of  compression savings thus calculates percentual data size decrease resulting from \emph{I-frame} omission as follows $100 *I_{size} / (I_{size} + P_{size})$, where $I_{size}$ and $P_{size}$ denote the file sizes of the \emph{I-frame} and \emph{P-frame}, respectively.\par 
%Consequently, the ratio of $I_{size}$ to the sum of $I_{size} + P_{size}$ signifies the extent of savings accomplished.\par
\subsection{Evaluation}
\begin{figure}[!ht]
  \centering
  \includegraphics[width=0.75\linewidth]{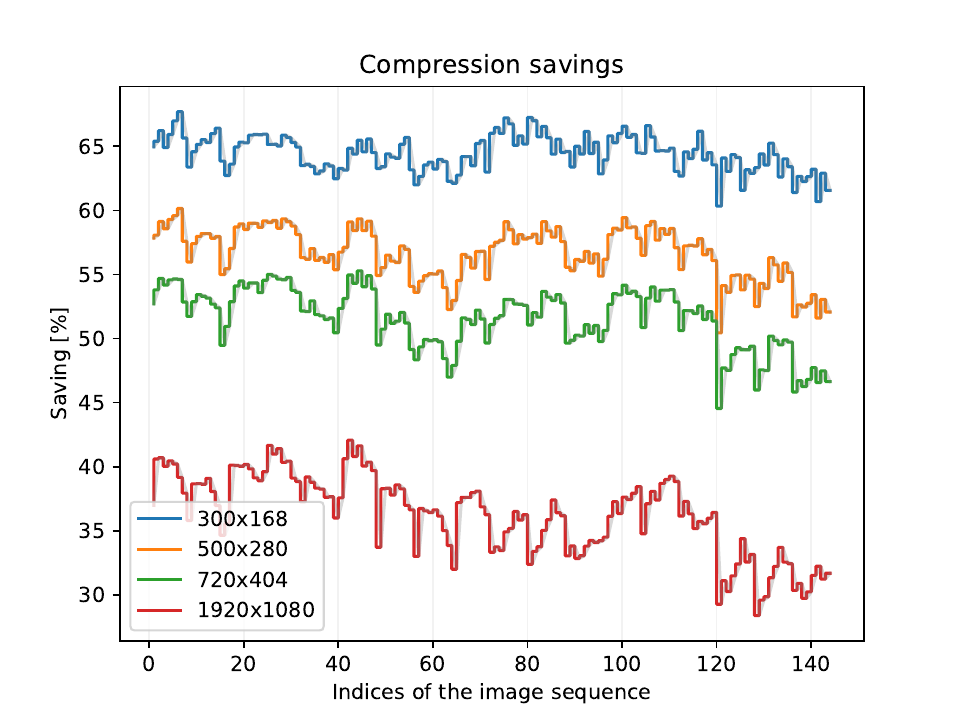}
   %{\epsfig{file = figures/hmap.pdf}}
  \caption{Encoder setting-averaged compression savings relative to H.264 achieved for individual images in empty scene scenario using INGP.}
\label{fig:step_plot_compression_savings_carla_ir_rl_noon_empty_ingp}
 \end{figure}

\begin{figure}[!ht]
  \centering
  \includegraphics[width=0.75\linewidth]{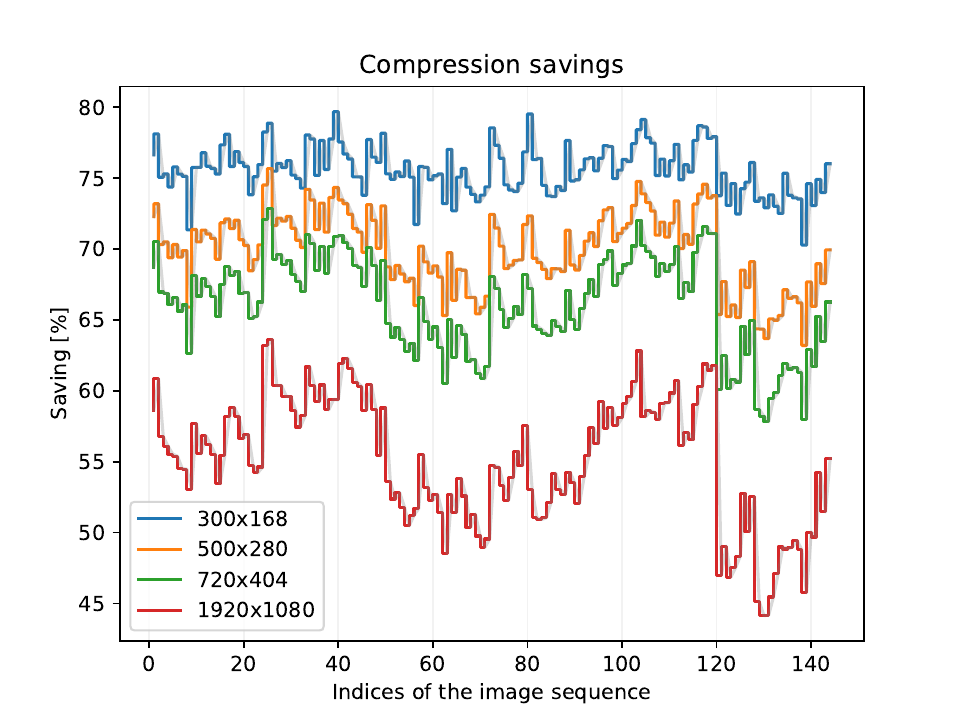}
   %{\epsfig{file = figures/hmap.pdf}}
  \caption{Encoder setting-averaged compression savings relative to H.264 achieved for individual images in empty scene scenario using 3DGS.}
\label{fig:step_plot_compression_savings_carla_ir_rl_noon_empty_gs}
 \end{figure}

Table \ref{tab:ingp_3dgs_metrics} presents an evaluation of the RF models utilizing PSNR, SSIM, and LPIPS metrics to reveal differences between GT and RF-generated image. 
%This assessment was conducted on a dataset comprising 144 images, encompassing two distinct scenarios.
\begin{comment}
\begin{table}[]
\centering
\caption{Average mean values of RF models}
\label{tab:ingp_3dgs_metrics}
\begin{tabular}{|c|cc|cc|cc|}
\hline
\textbf{Metrics} & \multicolumn{2}{c|}{\textbf{PSNR} $\uparrow$} & \multicolumn{2}{c|}{\textbf{SSIM} $\uparrow$} & \multicolumn{2}{c|}{\textbf{LPIPS $\downarrow$}} \\ \hline
\textbf{Scenario} & \multicolumn{1}{c|}{INGP} & 3DGS & \multicolumn{1}{c|}{INGP} & 3DGS & \multicolumn{1}{c|}{INGP} & 3DGS \\ \hline
Empty & \multicolumn{1}{c|}{26.33} & 29.41 & \multicolumn{1}{c|}{0.75} & 0.85 & \multicolumn{1}{c|}{0.38} & 0.24 \\ \hline
Vehicles & \multicolumn{1}{c|}{21.33} & 21.78 & \multicolumn{1}{c|}{0.68} & 0.71 & \multicolumn{1}{c|}{0.44} & 0.30 \\ \hline
\end{tabular}
\end{table}
\end{comment}
\begin{table}[]
\centering
\caption{Average mean values of RF models}
\label{tab:ingp_3dgs_metrics}
\begin{tabular}{c|cc|cc|cc}
\textbf{Metrics} & \multicolumn{2}{c|}{\textbf{PSNR} $\uparrow$} & \multicolumn{2}{c|}{\textbf{SSIM} $\uparrow$} & \multicolumn{2}{c}{\textbf{LPIPS $\downarrow$}} \\ \hline
\textbf{Scenario} & \multicolumn{1}{c|}{INGP} & 3DGS & \multicolumn{1}{c|}{INGP} & 3DGS & \multicolumn{1}{c|}{INGP} & 3DGS \\ \hline
Empty & \multicolumn{1}{c|}{26.33} & 29.41 & \multicolumn{1}{c|}{0.75} & 0.85 & \multicolumn{1}{c|}{0.38} & 0.24 \\ \hline
Vehicles & \multicolumn{1}{c|}{21.33} & 21.78 & \multicolumn{1}{c|}{0.68} & 0.71 & \multicolumn{1}{c|}{0.44} & 0.30 \\
\end{tabular}
\end{table}
In the scenario devoid of external elements not present in RF, the INGP model exhibited notably inferior performance, primarily attributable to its less precise reconstruction of street lamp structures compared to the 3DGS model, as illustrated in Fig.~\ref{fig:carla_ir_rl_ingp_gs_gt2}. This disparity in model efficacy is further reflected in the context of video compression savings. Fig. \ref{fig:step_plot_compression_savings_carla_ir_rl_noon_empty_ingp} and Fig. \ref{fig:step_plot_compression_savings_carla_ir_rl_noon_empty_gs} demonstrate that compression savings of the 3DGS model are ranging from 45\% to 80\%, whereas the INGP model achieved a lower range of 30\% to 68\%.

\begin{figure}[!ht]
  \centering
  \includegraphics[width=0.76\linewidth]{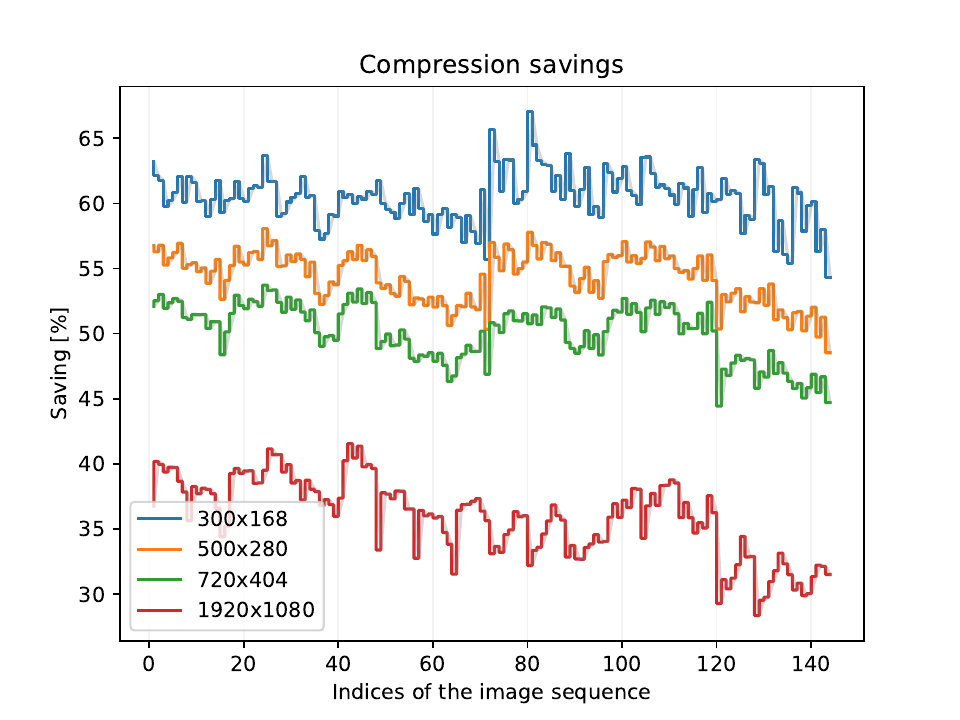}
   %{\epsfig{file = figures/hmap.pdf}}
  \caption{Encoder setting-averaged compression savings relative to H.264 achieved for individual images in vehicles in the scene scenario using INGP.}
\label{fig:step_plot_compression_savings_carla_ir_rl_noon_vehicles_ingp}
 \end{figure}

\begin{figure}[!ht]
  \centering
  \includegraphics[width=0.76\linewidth]{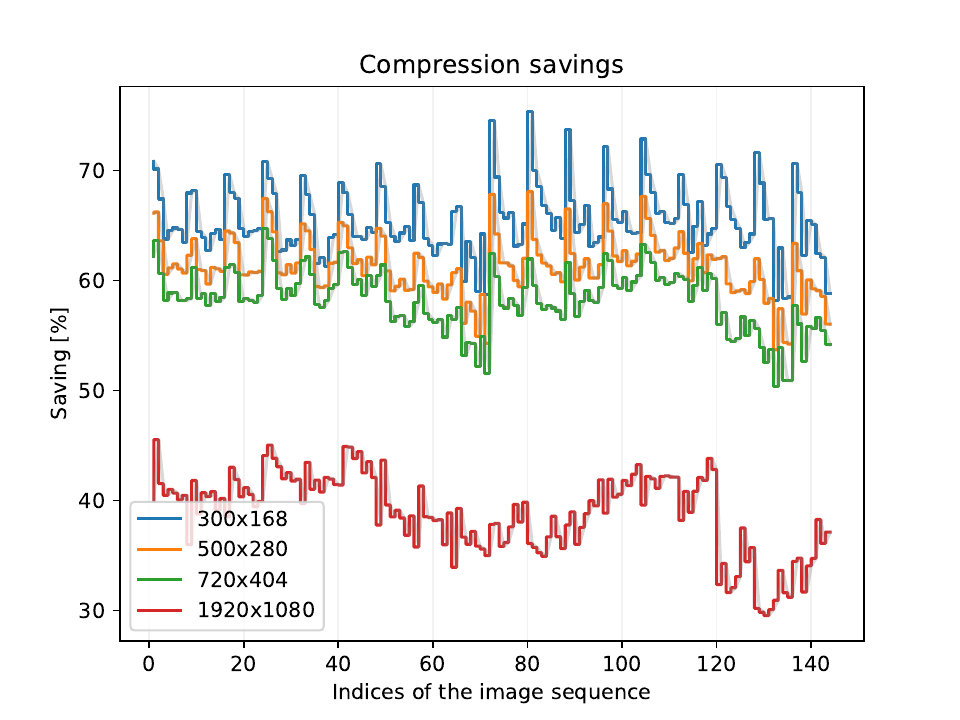}
   %{\epsfig{file = figures/hmap.pdf}}
  \caption{Encoder setting-averaged compression savings relative to H.264 achieved for individual images in vehicles in the scene scenario using 3DGS.}
\label{fig:step_plot_compression_savings_carla_ir_rl_noon_vehicles_gs}
 \end{figure}

Conversely, in the scenario incorporating vehicles into the 3D scene, while using RFs in which they were absent, both RF models experienced a decline in metric performance due to the introduction of these additional objects. The differences in performance between the models in this scenario were more nuanced. As per Table~\ref{tab:ingp_3dgs_metrics}, the LPIPS metric was consistent with the empty scenario, showing similar variance. 
However, Fig.~\ref{fig:step_plot_compression_savings_carla_ir_rl_noon_vehicles_ingp} and Fig.~\ref{fig:step_plot_compression_savings_carla_ir_rl_noon_vehicles_gs} suggest more subtle distinctions between the models in this scenario compared to the former. In this case, the INGP model exhibited compression savings between 28\% and 68\%, while the 3DGS model ranged from 30\% to 75\%.
From the analysis of the scenario with vehicles, it is evident that there is a convergence in both evaluation metrics and compression savings between the models. This suggests that increased variability in the real scene leads to more homogeneous evaluation results of the RF-generated images, indicating a correlation between scene complexity and model performance consistency. \par

 Additional outcome is the observed inverse relationship between resolution and compression savings, partly due to the imperfections in images generated by RF models. In the downscaling process, high-frequency signal components are lost in both original and RF-generated images, acting as a form of lossy compression. This loss brings RF-generated images closer to the GT by removing high-frequency artifacts, highlighting the complex interplay between resolution and compression effectiveness.

\section{Conclusion}
In this work, we propose a compression approach using RF-based encoder and decoder, that can simultaneously serve as a distributed photorealistic metaverse backbone, with low real-to-sim synchronization latency.
We have first evaluated feasibility of such approach in review of related work, and then experimentally on a dataset of camera images and videos captured by simulated CAVs in CARLA simulator. Our results show that our approach can achieve significant data compression and high reconstruction quality. We believe that our approach can pave the way for more scalable and  realistic metaverse services for CAVs and beyond.

\section*{Acknowledgment}
This work was supported by The Slovak Research and Development Agency project no. APVV SK-CZ-RD-21-0028 and the Slovak Academy of Sciences project no. VEGA 1/0685/23.

\bibliography{bibliography}{}
\bibliographystyle{IEEEtran}
\end{document}